\documentclass[10pt,twocolumn,letterpaper]{article}

\usepackage{cvpr}  
\usepackage{multirow}
\usepackage{url}
\usepackage{graphicx}
\usepackage{booktabs}
\usepackage{animate}
\usepackage[misc]{ifsym}
\usepackage{tabularx}
\usepackage{subcaption}
\usepackage{colortbl}
\usepackage{tabularx,booktabs}

\usepackage[dvipsnames]{xcolor}

\definecolor{cvprblue}{rgb}{0.21,0.49,0.74}
\usepackage[pagebackref,breaklinks,colorlinks,citecolor=cvprblue]{hyperref}

\newcommand{\name}{MoonShot}
\newcommand{\blockname}{multimodal video block}

\def\paperID{6298} 
\def\confName{CVPR}
\def\confYear{2024}

\title{\name: Towards Controllable Video Generation and Editing with Multimodal Conditions}

\author{
  David Junhao Zhang$^{\blacklozenge}$$^{\diamondsuit}$,
  Dongxu Li$^{\blacklozenge}$,
  Hung Le$^{\blacklozenge}$,
  Mike Zheng Shou$^{\diamondsuit}$\protect\thanks{Corresponding author},
  Caiming Xiong$^{\blacklozenge}$,
  Doyen Sahoo$^{\blacklozenge}$ \\
  \\
  $^{\blacklozenge}$Salesforce Research \quad \quad
  $^{\diamondsuit}$Show Lab, National University of Singapore
  \\
\url{https://showlab.github.io/Moonshot/}
  \\}

\begin{document}

\twocolumn[{%

\maketitle
\renewcommand\twocolumn[1][]{#1}%
\setlength{\tabcolsep}{0.5pt}
\renewcommand{\arraystretch}{0.5}
\vspace{-9mm}
 \includegraphics[width=1.0\textwidth]{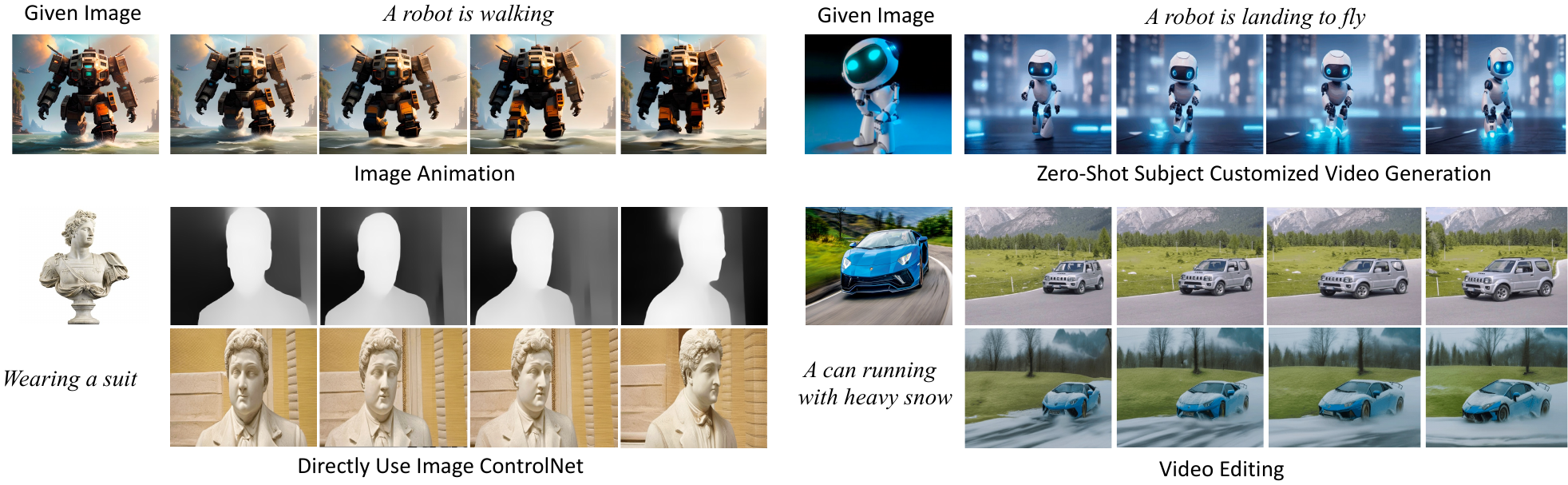}
 
\captionof{figure}{Our foundational video diffusion model, underpinned by multimodal conditioning, effectively facilitates image animation, the creation of customized videos, and precise control over geometric structures through the image ControlNet. Additionally, it enables video editing guided by text and image inputs, producing better results than recent state-of-the-art methods in these applications.}
\vspace{3mm}

\label{fig:teaser}
}]



\footnote{$^{*}$Corresponding Author}
\begin{abstract}

Most existing video diffusion models (VDMs) are limited to mere text conditions. Thereby, they are usually lacking in control over visual appearance and geometry structure of the generated videos. This work presents~\name, a new video generation model that conditions simultaneously on multimodal inputs of image and text. The model builts upon a core module, called multimodal video block (MVB), which consists of conventional spatialtemporal layers for representing video features, and a decoupled cross-attention layer to address image and text inputs for appearance conditioning. In addition, we carefully design the model architecture such that it can optionally integrate with pre-trained image ControlNet modules for geometry visual conditions, without needing of extra training overhead as opposed to prior methods. Experiments show that with versatile multimodal conditioning mechanisms, \name~demonstrates significant improvement on visual quality and temporal consistency compared to existing models. In addition, the model can be easily repurposed for a variety of generative applications, such as personalized video generation, image animation and video editing, unveiling its potential to serve as a fundamental architecture for controllable video generation. Models will be made public on \url{https://github.com/salesforce/LAVIS}.
\end{abstract}
    
\vspace{-2mm}
\section{Introduction}
\label{sec:intro}

Recently, text-to-video diffusion models (VDMs)~\cite{singer2022make, ho2022imagen, blattmann2023align, esser2023structure, he2022latent, wang2023modelscope, ge2023preserve, zhou2022magicvideo, zhang2023show,zhao2023motiondirector}
have developed significantly, allowing creation of high-quality visual appealing videos.
However, most existing VDMs are limited to mere text conditional control, which is not always sufficient to precisely describe visual content.
Specifically, these methods are usually lacking in control over the \emph{visual appearance} and \emph{geometry structure} of the generated videos, rendering video generation largely reliant on chance or randomness.


%

%

%

\begin{figure}[t]
  \centering
 
   \includegraphics[width=1.0\linewidth]{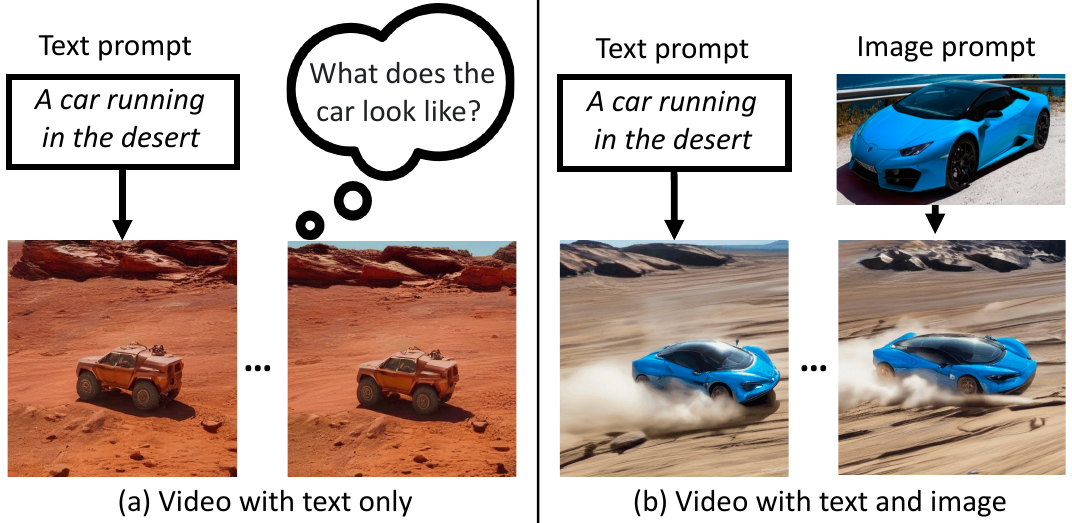}
\vspace{-1mm}
   \caption{A  single text prompt (a) lacks the precision and detail for accurate subject description. However, a picture (b) conveys much more, worth a thousand words. By combining a picture with text, we can produce videos that more closely match user requirements.}
   \label{fig:one}
\end{figure}

It is well acknowledged that text prompts are not sufficient to describe precisely the appearance of generations, as illustrated in Fig.~\ref{fig:one}.
To address this issue, in the context of text-to-image generation, efforts are made to achieve personalized generation~\cite{dreambooth,customdiffusion,blipdiffusion,adapter,photoverse} by fine-tuning diffusion models on input images.
Similarly for video generation, AnimateDiff relies on customized model weights to inject conditional visual content, either via LoRA~\citep{lora} or DreamBooth tuning~\citep{dreambooth}.
Nonetheless, such an approach incurs repetitive and tedious fine-tuning for each individual visual conditional inputs, hindering it from efficiently scaling to wider applications.
This inefficiency, as hinted by the prior work~\citep{blipdiffusion}, stems from the fact that most pre-trained text-to-video models are not able to condition both on images and text inputs.
To overcome this issue, we introduce a \emph{decoupled multimodal cross-attention} module to simultaneously condition the generation on both image and text inputs, thus facilitating to better control the visual appearance, while minimizing required fine-tuning efforts and unlocking zero-shot subject customized video generation.

In terms of geometric structure control, despite methods such as ControlNet~\citep{controlent} and T2I-Adapter~\citep{mou2023t2i} are developed to leverage depth, edge maps as visual conditions for image generation, analogous strategies for video synthesis remain indeterminate.
Among the few existing attempts, 
%
VideoComposer~\citep{videocomposer} adds video-specific ControlNet modules to video diffusion models (VDMs) and subsequently re-trains the added modules from scratch, incurring substantial extra training overhead. 
\begin{figure}[t]
  \centering
\includegraphics[width=0.95\linewidth]{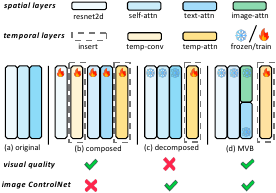}
\vspace{-1mm}
   \caption{Different designs of spatial-temporal modules include: (a) the original spatial module from U-Net, (b) a temporal module added within the spatial module, which hinders the image controlnet, and (c) a temporal module appended after the spatial module, allowing for image control network functionality but failing to produce high-quality videos with text-only conditioning. In contrast, our MVB block, conditioned on both image and text, enables the image controlnet and can generate high-quality videos.}
   \label{fig:two}
\end{figure}
In contrast, alternative methods~\cite{controlavideo,controlvideo} reuse pre-trained ControlNet modules for images.
However, they require adapting text-to-image models for video generation via frame propagation~\cite{yang2023rerender,tokenflow2023} or cross-frame attention~\cite{wu2022tune,text2video-zero}, resulting in subpar temporal consistency compared to those based on VDMs.

Building upon the aforementioned observations, our objective is to explore a model architecture that combines the strengths of both realms.
Namely, we expect the model to be a high quality VDM that produces consistent video frames, while also able to directly leverage pre-trained image ControlNet to condition on geometry visual inputs.
%
%
In this regard, we observe that as shown in Fig.~\ref{fig:two}(b), prior work~\citep{zhou2022magicvideo,gafni2022make,wu2022tune,wang2023modelscope} typically insert temporal modules, such as temporal convolutions layers, in between of spatial modules, usually before self-attention and after spatial convolution layers.
This design modifies the spatial feature distribution, thereby making direct integration with image ControlNet not feasible.
%
%
In contrast, AnimateDiff~\citep{animatediff} discards temporal convolution layers and only inserts temporal attention layers after spatial layers, as shown in Fig.~\ref{fig:two}(c).
Such design preserves spatial feature distribution thus facilitating immediate reuse of image ControlNet.
However, it depends exclusively on text conditions, which do not offer sufficient visual cues. Consequently, temporal modules have to learn extra spatial information for compensation, which diminishes the focus on temporal consistency, causing increased flickering and a decline in video quality.


To address these issues, we introduce \name, a video generation model that consumes both image and text conditional inputs.
The foundation of the model is a new backbone module for video generation, called \blockname~(MVB).
Specifically, each MVB highlights three main design considerations:
\begin{itemize}
    \item \emph{a conventional spatial-temporal module} for video generation, which in order consists of a spatial convolution layer, a self-attention layer and a temporal attention layer that aggregates spatial features. Such a design allows reuse of pre-trained weights from text-to-image generation models without altering its spatial feature distribution, thus subsuming its generation quality.
    \item \emph{a decoupled multimodal cross-attention layer} that conditions the generation on both text and image inputs. These two conditions complement each other to guide the generation. In addition, image input offers reference visual cues, allowing temporal modules to focus on video consistency. This improves overall generation quality and frame coherence, as evidenced experimentally;
    \item optionally, since spatial feature distribution is preserved, \emph{pre-trained image ControlNet modules} can be immediately integrated to control the geometric structure of the generation, without needing of extra training overhead.
\end{itemize}

As a result, our model generates highly consistent videos based on multimodal inputs, and can further utilize geometry inputs, such as depth and edge maps, to control the compositional layout of the generation.
Moreover, thanks to its generic architecture with versatile conditioning mechanisms, we show that \name~can be easily repurposed for a variety of generative applications, such as \emph{image animation} and \emph{video editing}.
Qualitative and quantitative results show that \name~obtains superior performance on personalized video generation, image animation and video editing.
When provided with a video frame as the image condition, \name~demonstrates competitive or better results with state-of-the-art foundation VDMs, validating the effectiveness of the model.




\begin{figure*}[t]
  \centering
 
   \includegraphics[width=0.8\linewidth]{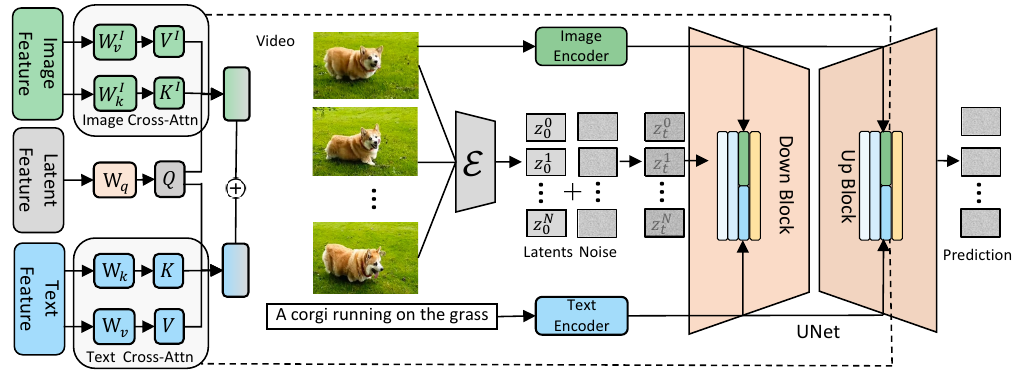}
\vspace{-2mm}
   \caption{The overall workflow and the structure of our decoupled multimodal cross-attention layers. In the training phase, we use the initial frame of the video as the image condition. For inference, the model accepts any image along with accompanying text.}
   \label{fig:three}
\end{figure*}

\begin{figure}[t]
  \centering
 
   \includegraphics[width=0.8\linewidth]{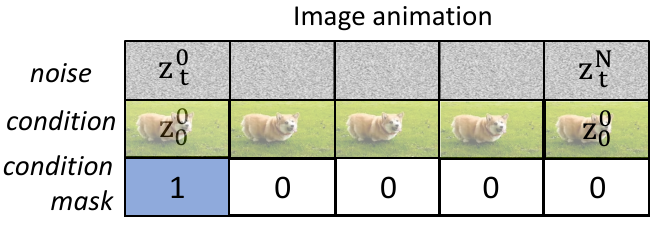}
\vspace{-2mm}
   \caption{Masked condition for image animation.}
   \label{fig:four}
\end{figure}

\section{Related Work}
\noindent\textbf{Text to Video Generation.}
Previous research in this field has leveraged a diversity of generative models, such as GANs~\cite{vondrick2016generating,saito2017temporal,Tulyakov_2018_CVPR,tian2021a,Shen_2023_CVPR}, Autoregressive models~\cite{srivastava2015unsupervised,yan2021videogpt,le2021ccvs,ge2022long,hong2022cogvideo}, and implicit neural representations~\cite{skorokhodov2021stylegan,yu2021generating}. Driven by the diffusion model's significant achievements in image synthesis, a number of recent studies have explored the application of diffusion models in both conditional and unconditional video synthesis~\cite{voleti2022masked,harvey2022flexible,zhou2022magicvideo,wu2022tune,blattmann2023videoldm,khachatryan2023text2video,hoppe2022diffusion,voleti2022masked,yang2022diffusion,nikankin2022sinfusion,luo2023videofusion,an2023latent,wang2023videofactory, wu2022nuwa}. 
Most existing approaches focus on conditioning a single modality. For instance, Imagen Video~\cite{ho2022imagen} and Make-A-Video are conditioned exclusively on text, while I2VGen-XL~\cite{2023i2vgenxl} relies solely on images.

In contrast, our method supports multimodal conditions with both image and text, enabling more precise control. 

\noindent\textbf{Video Editing and ControlNet.} 
Beginning with  Tune-A-Video~\cite{wu2022tune}, numerous works~\cite{gu2023videoswap,qi2023fatezero,ceylan2023pix2video,text2video-zero,yang2023rerender,vid2vid-zero,tokenflow2023,chai2023stablevideo} adapt the Text to Image (T2I) Model~\cite{rombach2022high} for video editing. These methods introduce additional mechanisms like cross-frame attention, frame propagation, or token flow to maintain temporal consistency. While Dreammix~\cite{molad2023dreamix}  uses the VDM for video editing, it necessitates source video fine-tuning. In contrast, our method directly employs VDM for video editing, achieving temporally consistent videos without the need for fine-tuning or complex designs.

ControlNet~\citep{controlent} and T2I-Adapter~\citep{mou2023t2i} utilize structural signals for image generation. While ControlVideo~\cite{controlvideo} applies this approach with a T2I model for video generation, resulting in limited temporal consistency. Control-A-Video~\cite{controlavideo} integrates a temporal layer into both the base VDM and ControlNet, training them jointly. Gen-1~\cite{esser2023structure} concatenates structural signals with noise as the UNet input. However, without these signals, the base VDMs in Control-A-Video and Gen-1 will fail. Videocomposer~\cite{videocomposer} incurs high training costs by adding a video controlnet after VDM training. Our method stands out as our VDM independently produces high-quality videos and directlu uses image ControlNets without extra training.

\noindent\textbf{Generation Model Customization.} Customizing large pre-trained foundation models~\cite{dreambooth, kumari2023multi, gu2023mix, chen2023anydoor, wei2023elite, smith2023continual} enhances user-specific preferences while retaining robust generative capabilities. Dreambooth~\cite{dreambooth} leads the way in this field, though it requires extensive fine-tuning. IP adapter~\cite{ye2023ip-adapter} and BLIP-Diffusion~\cite{blipdiffusion} achieve  zero-shot customization using  an additional image cross-attention layers and text-aligned image embeddings. Animatediff~\cite{animatediff} pioneers subject customization in videos but needs additional Dreambooth training for the new subject.

Our method differs from IP-adapter in image domains by introducing a decoupled image-text cross-attention layer in the video block. 
We explore its effectiveness in facilitating high-quality and smooth video creation and further apply it to video tasks involving image animation and video editing. Unlike Animatediff, our approach enables zero-shot subject customization directly in videos.

\noindent\textbf{Image Animation.} 
This task aims to generate subsequent frames given an initial image. Existing methods~\cite{i2v_3,i2v_36,i2v_41,i2v_53,i2v_81,i2v_41,i2v_36,ni2023conditional} have focused on specific domains like human actions and nature scenes, while recent advancements like I2VGen-XL~\cite{2023i2vgenxl}, DynamicCrafter~\cite{xing2023dynamicrafter}, and Videocomposer~\cite{videocomposer} target open-domain image animation with video diffusion models. However, they often struggle to maintain the appearance of the conditioning image.

In contrast, our approach, using masked condition and decoupled image-text attention, effectively aligns the first frame of the animated video with the provided image, ensuring a more accurate retention of its original identity. Our work is concurrent with DynamicCrafter~\cite{xing2023dynamicrafter} (Oct.,2023) and we make comparisons in Fig.~\ref{fig:i2v}.


\section{Model Architecture and Adaptations}
Text-to-video latent diffusion models generate videos by denoising a sequence of Gaussian noises with the guidance of text prompts.
The denoising network $\theta$ is usually a U-Net-like model, 
%
optimized by a noise prediction loss:
\begin{equation}
    \mathcal{L} = \mathbb{E}_{z_0, y, \epsilon\sim\mathcal{N}(0, \mathit{I}), t\sim\mathcal{U}(0, \mathit{T})}\left\lbrack 
\lVert \epsilon - \epsilon_\theta(z_t, t, y)) \rVert_2^2 \right\rbrack,
\label{euqation: noise_preditcion_loss}
\end{equation}
where $z_0$ is the latent code of training videos from VAE encoder, $y$ is the text prompt, $\epsilon$ is the Gaussian noise added to the latent code, $t$ is the time step and $\epsilon_{\theta}$ is the noise prediction by the model.
In the following, we first introduce the \name~model architecture including its core component, multimodal video block.
Then, we showcase the capabilities of the model by describing methods to repurpose the model for various video generation tasks, such as geometry controlled video generation, image animation and video editing.

\subsection{Multimodal Video Block}
Our model architecture builds upon multimodal video blocks (MVB).
There are three primary objectives underlying the design of this key module.
First, we aim for the model to consistently produce video frames of high quality.
Second, it is desired for immediate integration of pre-trained image ControlNet. In this way, we can facilitate the use of geometric images for controlling the compositional layout without extra training.
Third, the model is expected to accommodate multimodal text and image inputs for better visual appearance conditioning.
To this end, each MVB consists of two groups of layers, spatialtemporal U-Net layers and decoupled multimodal cross-attention layers as detailed in order below.

\noindent\textbf{Spatialtemporal U-Net Layers.}
Typical U-Net in text-to-image models consists in order of a spatial convolution layer (ResNet2D), a self-attention layer, and a cross-attention layer that conditions the generation on texts.
Prior work~\cite{singer2022make, blattmann2023align, wang2023modelscope} adapts these models for video generation by introducing additional temporal convolution and attention layers.
As shown in Fig.~\ref{fig:two}(b), the temporal convolution layer is usually inserted before each self-attention layer and after the spatial convolution layer.
While this architecture may enhance temporal consistency, it alters the distribution of the spatial feature baked in the pre-trained text-to-image generation models.
As a result, the model not only loses the ability of text-to-image generation, but also becomes incompatible with established techniques developed for text-to-image models, such as ControlNet, making direct integration of these techniques infeasible.

Differently, we observe that the addition of temporal attention layers after the cross-attention layer does not significantly modify the spatial feature distribution, while contributing effectively for temporal feature aggregation.
By freezing the spatial layers during training, we can reuse ControlNet to condition the generation on geometry visual inputs by broadcasting it along the temporal axis, as shown in Fig.~\ref{fig:two}(c).
In particular, we use space-time attention similar to~\citep{timesformer}, where each patch attends to those at the same spatial location and across frames.
\noindent\textbf{Decoupled Multimodal Cross-attention Layers.}
Most existing video generation models use a cross-attention module to condition the generation on texts.
Given $f_y$ the embedding of text prompts, diffusion models condition on it to enhance the U-Net features $f_x$ via cross-attention layers, where the query $Q$
is obtained from U-Net features $f_x$, key $K$ and value $V$ are from the text embedding $f_y$. The
cross-attention operation is then defined as:
\begin{equation}
\left \{
\begin{array}{ll}
    \mathbf{Q} = \mathbf{W}_{Q} \cdot f_x;\ \mathbf{K} = \mathbf{W}_{K}\cdot f_y;\ \mathbf{V} = \mathbf{W}_{V}\cdot f_y; \\
    \text{CrossAttention}(\mathbf{Q}, \mathbf{K}, \mathbf{V}) = \text{softmax}(\frac{\mathbf{Q}\mathbf{K}^T}{\sqrt{d}})\cdot \mathbf{V},
\end{array}
\right.
\label{eq:ldm_ca}
\end{equation}
where $Q \in \mathbb{R}^{BN\times H\times W \times C}$, $K,V\in \mathbb{R}^{BN \times L \times C}$, with $B$ the batch size, $N$ the number of frames, $H$ the height, $W$ the width and $C$ the number of channels, $L$ the number of text tokens, $d$ the hidden size. Note that text embeddings are duplicated for video frames.



There are two issues with this design. First, relying solely on text prompts usually proves inadequate in accurately describing highly customized visual concepts for the desired generation.
Second, specifically for video generation, the absence of visual conditioning mechanism places excessive burden on the temporal attention layers.
They must simultaneously ensure the consistency across frames and also preserve high-quality spatial features, often resulting in compromises on both fronts with flickered low-quality videos.

To address these issues, we introduce decoupled multimodal cross-attention, where one extra key and value transformation are optimized for image conditions, denoted as $K^{I}, V^{I}\in \mathbb{R}^{BN \times L \times C}$.
The attention is formulated as:
\begin{equation}
    \text{CrossAttention}(\mathbf{Q}, \mathbf{K}, \mathbf{V}) + \text{CrossAttention}(\mathbf{Q}, \mathbf{K}^{I}, \mathbf{V}^{I}).
\end{equation}
This approach enables the model to effectively manage both the image and text conditions.
Additionally, conditioning on visual cues allows the subsequent temporal modules to focus more on maintaining temporal consistency, resulting in smoother and higher-quality video outputs. With the additional image condition, the training loss is thereby reformulated as:
\begin{equation}
    \mathcal{L} = \mathbb{E}_{z_0, y, y{'}, \epsilon\sim\mathcal{N}(0, \mathit{I}), t\sim\mathcal{U}(0, \mathit{T})}\left\lbrack 
\lVert \epsilon - \epsilon_\theta(z_t, t, y,y')) \rVert_2^2 \right\rbrack,
\label{euqation: noise_preditcion_loss}
\end{equation}
where $y{'}$ is the image condition.

\subsection{Adapting for Video Generation Applications}

\noindent\textbf{Masked Condition for Image Animation.}
With the additional conditioning on image input, our model is a natural fit for the task of image animation, which aims to transform an input image into a short video clip with consistent content.
To enhance the content consistency, we adapt the mask-conditioning mechanism  introduced by \cite{blattmann2023align} for image animation. In particular, we use the first frame as an additional input condition for the U-Net. Apart from the original four latent channels, as illustrated in Fig.~\ref{fig:four}, we add five more channels to the U-Net's input. Among them, four channels represent the replicated first frame latent $z_{0}^{0}$, and one binary channel is used to denote the masked frames. This approach encourages that the identity of the subject in the animated video remains identical to that in the conditioning image.
We observe that incorporating an extra image cross attention layer is essential for image animation.
It helps significantly to prevent sudden changes in appearance and reduce temporal flickering, which are seen commonly in models driven merely by text conditions.

\noindent\textbf{Video Editing with Video Diffusion Models.}
DreamMix~\cite{molad2023dreamix} shows that VDM can be repurposed for video editing, which however requires extensive fine-tuning.
Our model is for general-purpose video generation, yet can be used for video editing without needing of fine-tuning.
Specifically, for a selected encoded source video $z_{0}$, we add Gaussian noise using the DDPM~\cite{ho2020denoising} forward process. Next, we employ diffusion directly with VDM, conditioned on both text and image. This process effectively replaces the subject in the original video with that from the image condition and incorporates visually appealing elements as described in the text, resulting in a smooth edited video.

\noindent\textbf{Geometry Controlled Generation.}
Since our model preserves the spatial features of the pre-trained text-to-image models, we can directly integrate the image ControlNet for geometry conditioning. 
To achieve this, we attach pre-trained ControlNet modules to the model.
Then condition features for each frame is added to the corresponding feature maps via residues.
Due to the careful design of spatialtemporal U-Net layers, we observe satisfactory geometry control effect without needing of video-specific fine-tuning.

\section{Experiments}

\begin{figure}[t]
  \centering
 
   \includegraphics[width=1.0\linewidth]{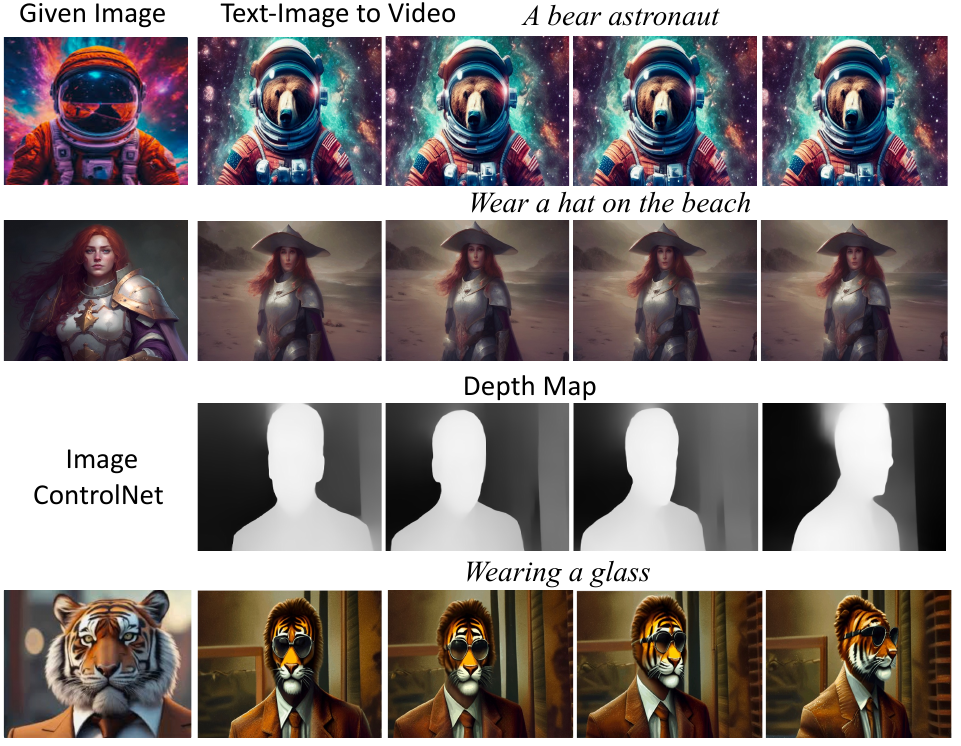}
\vspace{-6mm}
   \caption{Subject customized video generation results with the option of utilizing an image controlnet network or not.}
   \label{fig:subject}
\end{figure}

\subsection{Implementation Details}

Our spatial weights are initialized using SDXL~\cite{podell2023sdxl}, which are fixed throughout the training process. Initially, following IP-Adpter~\cite{ye2023ip-adapter}, we train the image cross-attention layers using the LAION~\cite{laion5b} dataset at a resolution of $512\times 320$. Subsequently, we keep the spatial weights unchanged and proceed to train only the temporal attention layers. This training step utilizes the WebVid10M~\cite{bain2021frozen} dataset, each clip sampled 16 frames at a $512\times 320$
 resolution, with conditioning on video captions as text conditions and the first frame as image conditions. Further refinement is carried out on a set of 1000 videos from the InternVideo~\cite{wang2022internvideo} collection, aimed at removing watermarks. We use 16 A100 40G GPUs for training. More details can be found in supplementary material.

\begin{figure*}[t]
  \centering
 
   \includegraphics[width=1\linewidth]{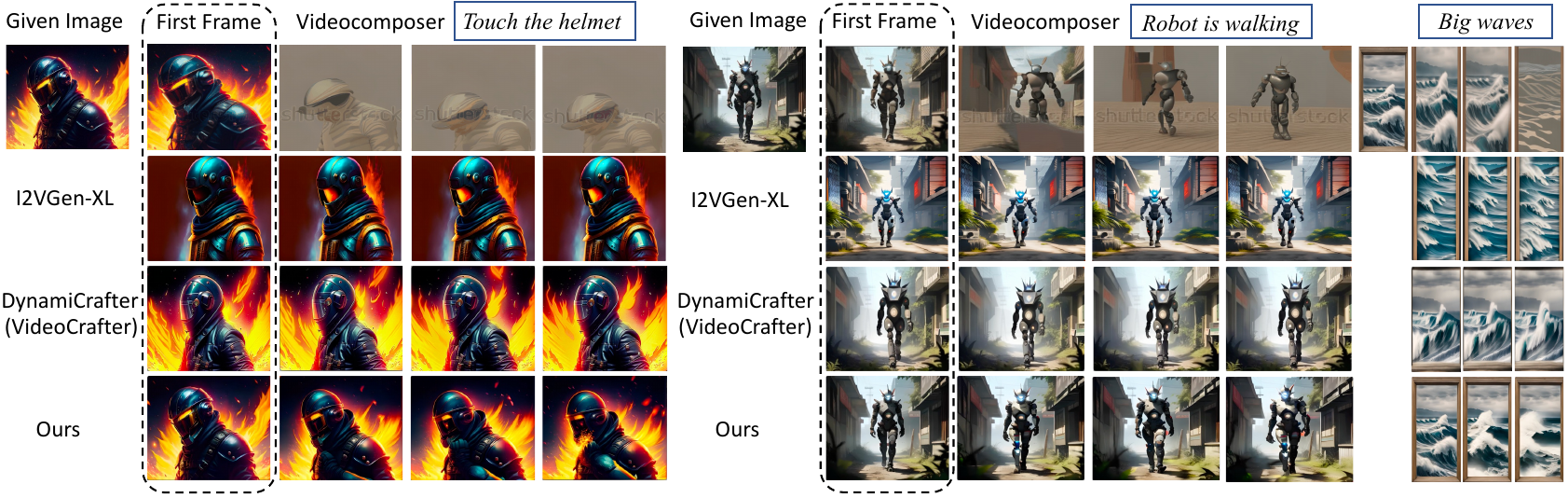}
\vspace{-6mm}
   \caption{Image animation. Comparing our method with I2VGEN-XL~\cite{2023i2vgenxl}, DynamiCrafter~\cite{xing2023dynamicrafter}, and VideoComposer~\cite{videocomposer}, it stands out in its ability to control the animated video's first frame to match the provided image exactly, maintaining the identity of the given image more precisely and animating the image according to text prompts. In contrast, I2VGen-XL and DynamicCrafter show relatively weaker identity preservation in their animated videos, and their response to text prompts is less effective.}
   \label{fig:i2v}
\end{figure*}

\subsection{Human Evaluations}
\label{human}
 We follow Make-a-Video~\cite{singer2022make} to perform human evaluations on Amazon Mechanical Turk. In the realm of video editing tasks (Tab.~\ref{table:editing}), in line with FateZero~\cite{qi2023fatezero} and Render-A-Video~\cite{yang2023rerender}, we direct annotators to identify the most superior outcomes from five methods, judging by three standards: 1) the precision of prompt-to-edited-frame alignment, 2) temporal coherence of the video, and 3) the overall quality. In addition, for ControlNet evaluations (see Tab.~\ref{table:ablation1}), we request annotators to judge whether the created video adheres to the control signal. Furthermore, regarding the text-to-video ablation studies (Tab.~\ref{table:ablation2}), we invite annotators to evaluate the overall video quality, the accuracy of text-video alignment and motion fidelity.

\subsection{Subject Customized Generation}

\begin{table}[t]
\centering
\resizebox{0.80\linewidth}{!}{
\begin{tabular}{cccc}
\toprule
Model & DINO & CLIP-I & CLIP-T \\
\hline
Non-Customized T2V &0.283 & 0.594 & \textbf{0.296}\\
I2VGen-XL~\cite{2023i2vgenxl} & 0.542 & 0.737 & 0.218 \\
\hline
AnimateDiff~\cite{animatediff}  & \multirow{2}{*}{0.582} & \multirow{2}{*}{0.784} & \multirow{2}{*}{0.243} \\
 300 finetune steps &   & & \\
 \hline
Ours (zero-shot) & 0.556 & 0.763 & 0.292 \\
\hline
Ours (80 finetune steps) & \textbf{0.624} & \textbf{0.802} & 0.292\\
\bottomrule
\end{tabular}}
\vspace{-1.25mm}
\caption{Subject customized video generation performance on Dreambooth dataset~\cite{dreambooth}. Higher metrics are better.}
\label{table:subject}
\end{table}

\noindent\textbf{Quantitative Results.} 
To evaluate our method for subject-customized video generation, we perform experiments on the DreamBooth~\cite{dreambooth} dataset, which includes 30 subjects, each with 4-7 text prompts. 
We utilize DINO and CLIP-I scores to assess subject alignment, and CLIP-T for video-text alignment, calculating average scores for all frames. As shown in Tab.~\ref{table:subject}, our method achieves strong zero-shot customization, surpassing non-customized text-to-video (T2V) models by a large margin. Different from AnimateDiff, which requires repetitive re-training for new subjects, our method utilizes pre-trained decoupled multimodal attention layers, achieving zero-shot customization with compared performance.
If fine-tuned with as few as 80 steps, our approach further surpasses AnimateDiff by a significant margin, demonstrating the effectiveness of our model.

\noindent\textbf{Qualitative Results.}  As shown in Fig.~\ref{fig:subject}, our model produces customized videos that align with both the subject of image condition and the text condition. Additionally, the image ControlNet can be directly integrated to realize control over geometric structures.

\begin{table}[t]
\centering

\resizebox{0.95\linewidth}{!}{
\begin{tabular}{lccc}

\toprule
& DINO (First) & DINO (Avg) & CLIP-T (Avg) \\
\hline
GT & 0.781 & 0.644 & -- \\
I2VGen-XL~\cite{2023i2vgenxl} & 0.624 & 0.573 &0.232 \\
VideoComposer~\cite{videocomposer} & 0.751 & 0.285 & 0.269 \\
Ours & \textbf{0.765} & \textbf{0.614} & \textbf{0.284} \\
\bottomrule
\end{tabular}}
\vspace{-2mm}
\caption{Image animation results. Our model shows better visual and textual alignment than competing methods.}
\label{table:i2v}

\end{table}

\subsection{Image Animation}
\noindent\textbf{Quantitative Results.} To assess image animation capabilities, we select 128 video-text pairs from the Webvid evaluation set, covering diverse themes. We use DINO (First) to measure the similarity between the first frame of the animated video and the conditioning image, DINO (Average) for the average similarity across all video frames compared to the conditioning image, and CLIP-T for the overall alignment between text prompts and the animated video across frames.
As shown in Tab.~\ref{table:i2v}, our method outperforms others in all metrics, demonstrating superior identity preservation, temporal consistence and text alignment.

\noindent\textbf{Qualitative Results.} 
We compare our results qualitatively with I2VGEN-XL~\cite{2023i2vgenxl}, DynamiCrafter~\cite{xing2023dynamicrafter}, and VideoComposer~\cite{videocomposer} as shown in Fig.~\ref{fig:i2v}, it's seen that the identity or appearance in animated videos from I2VGEN-XL and DynamiCrafter is different from the original image. While VideoComposer replicates the conditioning image as the first frame, subsequent frames show abrupt changes in appearance, as indicated by its high DINO (First) and low DINO (Average) scores in Tab.~\ref{table:i2v}. This issue may stem from its limited capacity to extract visual cues. Our method, in contrast, utilizes multimodal cross-attention layers and condition masks, and excels by promoting the similarity between the first frame of the animated video and the conditioning image, maintaining more effectively appearance, and enabling animation in line with text prompts.

\subsection{Video Editing}

\begin{figure}[t]
  \centering
 
   \includegraphics[width=1.0\linewidth]{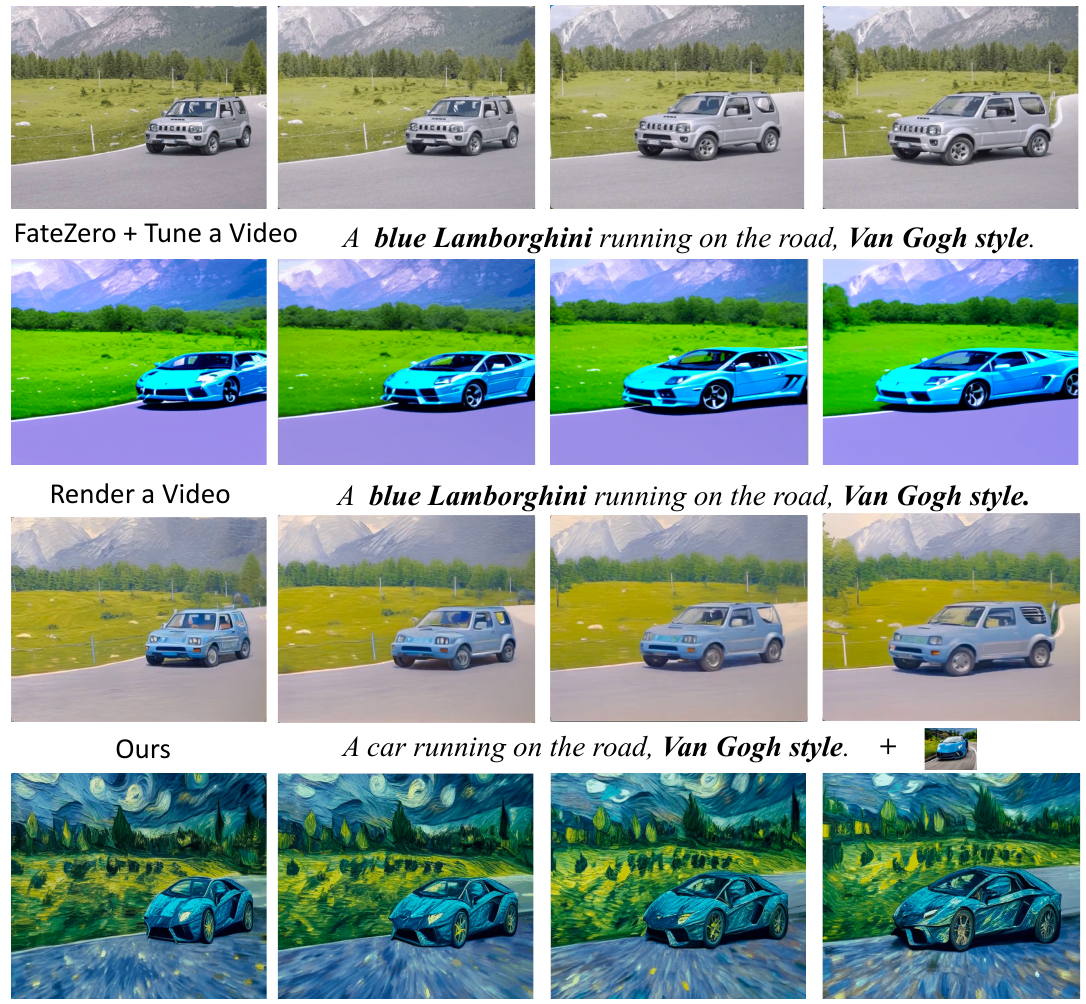}
\vspace{-4mm}
   \caption{Visual comparisons with SOTA video editing methods.}
   \label{fig:editing results}
\end{figure}

\begin{table}[t]
\centering

\resizebox{\linewidth}{!}{
\begin{tabular}{lccccc}
\toprule
Metric & FateZ~\cite{qi2023fatezero} & Pix2V~\cite{ceylan2023pix2video} & T2V-Z~\cite{text2video-zero} & Render-A-V~\cite{yang2023rerender} & Ours \\

\hline
Fram-Acc ($\uparrow$) & 0.534 & \textbf{0.978} & 0.943 & 0.959\%  &0.976\\
Tem-Con ($\uparrow$) & 0.953 & 0.942 & 0.963 & 0.965  &  \textbf{0.986} \\
Pixel-MSE ($\downarrow$) & 0.092 & 0.256 & 0.091 & 0.073 &\textbf{0.064} \\
\hline
User-Balance & 4.4\% & 6.2\% & 7.4\% & 21.4\%  & \textbf{60.6\%} \\
User-Temporal & 3.6\% & 2.0\% & 3.8\% & 18.2\%  &\textbf{72.4\%}\\
User-Overall & 3.1\% & 3.1\% & 7.0\% & 24.6\%  &\textbf{62.2\%} \\
\bottomrule
\end{tabular}}
\caption{Quantitative comparisons and user preference rates for video editing task.}
\label{table:editing}
\end{table}

\begin{figure*}[t]
  \centering
 
   \includegraphics[width=0.9\linewidth]{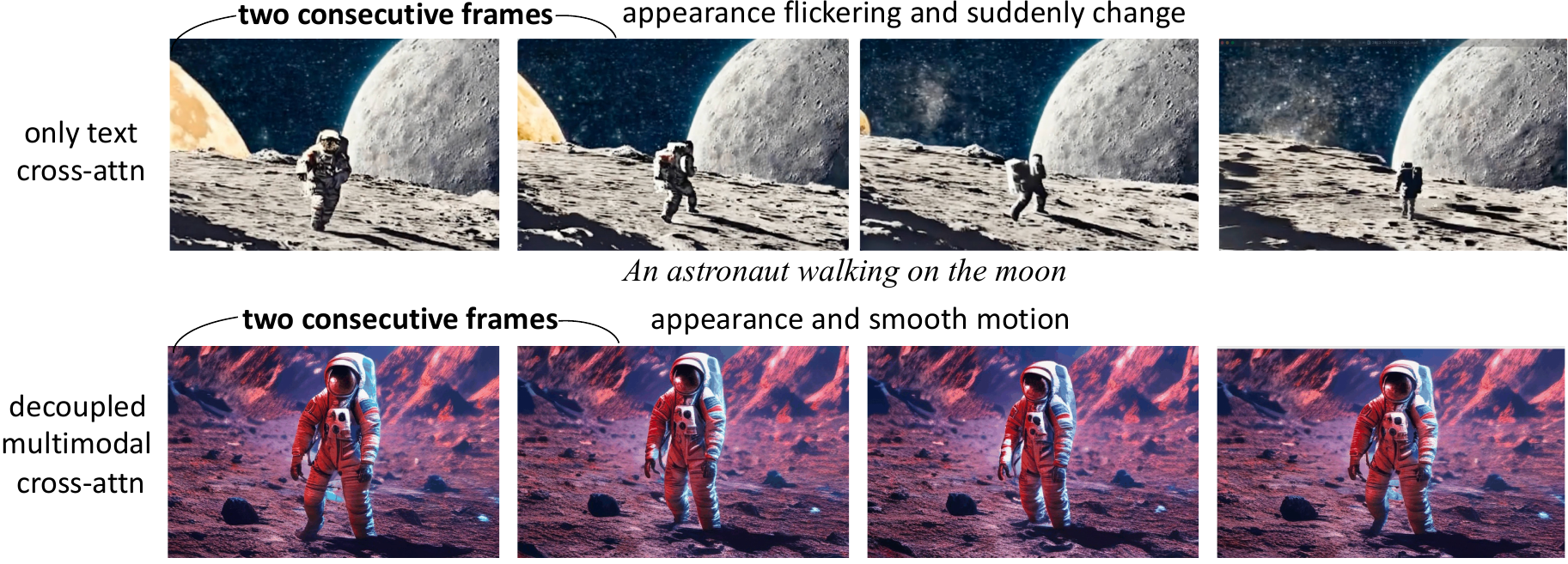}
   \caption{Comparison between text-only and multimodal conditioned VDMs. We compare \name~with AnimateDiff-XL~\cite{animatediff}, which uses only text conditions. Despite trained on high-quality internal data, AnimateDiff-XL suffers appearance flickering and abrupt changes. In contrast, our multimodal-conditioned model trained on a public dataset shows improved temporal consistency and visual quality.}
   \label{fig:ablation2}
\end{figure*}

\begin{figure}[t]
  \centering
 
   \includegraphics[width=0.9\linewidth]{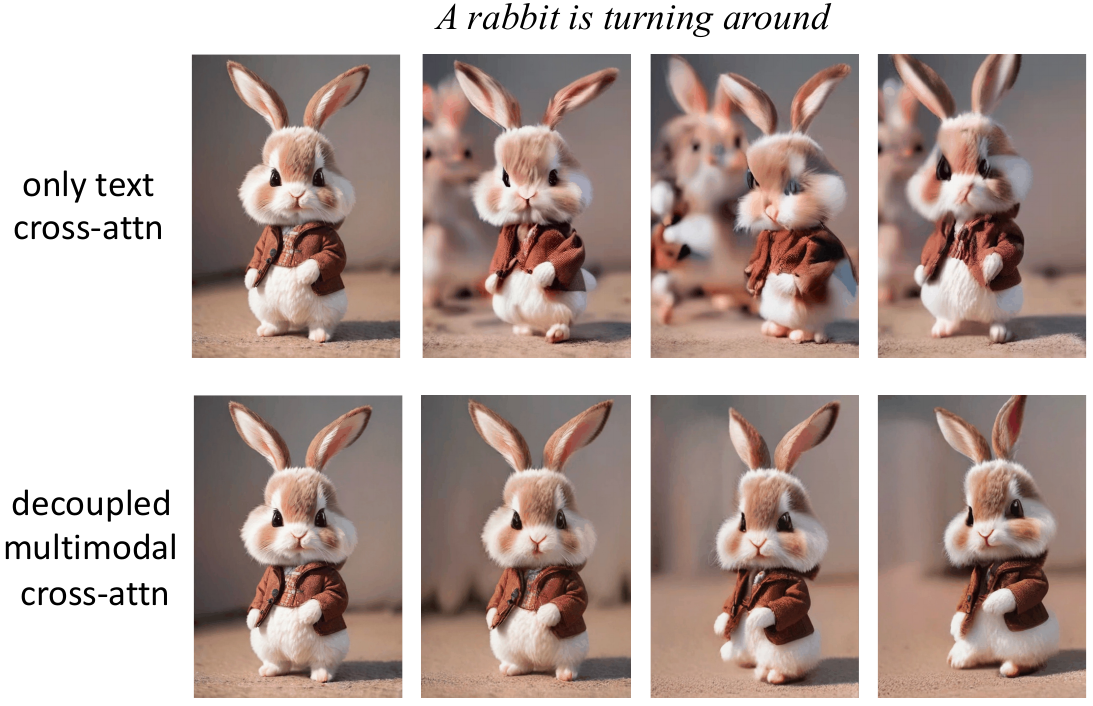}
   \caption{Impact of multimodal conditions for image animation.}
   \label{fig:ablation3}
\end{figure}

\noindent\textbf{Quantitative Results.} We compare with four video editing methods: FateZero~\cite{qi2023fatezero}, Pix2Video~\cite{ceylan2023pix2video}, Text2Video-Zero~\cite{text2video-zero}, and Render-A-Video~\cite{yang2023rerender}. Notably, Render-A-Video and Text2Video-Zero employ customized models that incorporate ControlNet. In contrast, our approach utilizes the base VDM model without integrating ControlNet. 
Following the FateZero and Pix2Video, we utilize 72 videos from Davis~\cite{pont20172017} and various in-the-wild sources. We report three metrics in Tab.~\ref{table:editing}: Fram-Acc, a CLIP-based measure of frame-wise editing accuracy; Tmp-Con, assessing the cosine similarity between consecutive frames using CLIP; and Pixel-MSE, the averaged mean-squared pixel error between aligned consecutive frames. Our method excels in temporal consistency and ranks second in frame editing accuracy. And our method achieves better human preference on all evaluation metrics (Sec.~\ref{human}).
In fact, prior methods typically utilize image models with frame propagation or cross-frame attention mechanisms, which tend to yield worse temporal consistency compared to our approach. 
This demonstrates the clear advantage of using foundation VDMs for video editing, compared to those relying on image models.


\noindent\textbf{Qualitative Results.} As shown in Fig.~\ref{fig:editing results}, FateZero reconstructs the input sports car frame well but the result is not aligned with the prompt. Render-A-Video struggles to swap the jeep to a sports car, facing challenges with shape changes. Conversely, our method adeptly replaces the jeep with a sports car as specified in the conditioning image while also adhering to the text prompts.

\subsection{Text to Video Generation}
\vspace{-2mm}
\begin{table}[t]
\centering

\resizebox{0.9\linewidth}{!}{
\begin{tabular}{c|ccc}
\toprule
Models              & FID-vid ($\downarrow$)   &  FVD ($\downarrow$) & CLIP-T ($\uparrow$) \\ \hline
N\"UWA~\citep{wu2022nuwa}       & 47.68   & -  & 0.2439        \\
CogVideo (Chinese)~\citep{hong2022cogvideo}   & 24.78 & - & 0.2614      \\
CogVideo (English)~\citep{hong2022cogvideo}   & 23.59  & 1294 & 0.2631  \\
MagicVideo~\citep{zhou2022magicvideo}          & -    & 1290 & -        \\
Video LDM~\citep{blattmann2023align}      & -   &  - & 0.2929 \\ 
Make-A-Video~\citep{singer2022make}        & 13.17 &  -& 0.3049     \\ 
ModelScopeT2V~\citep{wang2023modelscope}  &   11.09 & 550 & 0.2930    \\ 
\hline

Ours &  \textbf{10.98} & \textbf{542} & \textbf{0.3068}   \\
\bottomrule
\end{tabular}}

\caption{
Quantitative comparisons on MSR-VTT~\cite{xu2016msr}.}


\label{table:t2v}
\end{table}
Since the spatial layers are frozen during training, we first generate an image according to the text prompt. The image is later combined with text for multimodal conditioned generation.
%
We use MSR-VTT dataset~\cite{xu2016msr} to evaluate quality of zero-shot generation, whose test set contains 2,990 videos of 320x240 resolution, accompanied by 59,794 captions in total.
We compare our model with state-of-the-art methods as shown in Tab.~\ref{table:t2v}.
Our method achieves the best results across FID-vid~\cite{heusel2017gans_nash_equilibrium}, FVD~\cite{unterthiner2018FVD}, and CLIP-T~\cite{wu2021godiva}, demonstrating better visual quality and text alignment.

\subsection{Ablation Studies}
\noindent\textbf{Spatial temporal modules designs.}
We investigate the design of temporal modules that allows direct integration of the image ControlNet.
As indicated in Tab~\ref{table:ablation1}, the design which inserts the temporal convolution within spatial modules (Fig.\ref{fig:two}b) alters the original spatial features, rendering the image ControlNet ineffective, even with fixed spatial (fs) weights during  video training. Conversely, placing the temporal attention after all spatial modules in each block and fixing spatial weight during video training renders the image ControlNet feasible.

\noindent\textbf{Impact of image condition on video consistency and quality.} 
In Tab.~\ref{table:ablation2}, we explore the impact of mutlimodal condition on video generation. We freeze the spatial layers and only train the temporal ones.  Animatediff-XL also employs fixed SDXL spatial weights conditioned only by text and is included in this ablation. However, since it's trained on internal high-quality data, we've also trained a model without image conditions on the same data used for our final model, to ensure a fair ablation. Text-Video (T-V) Alignment and Motion Fidelity are agreement scores from human evaluation (Sec~\ref{human}). We find that using only text condition  leads to weaker temporal consistency, motion fidelity, and visual quality compared to multimodal conditions. Fig.~\ref{fig:ablation2} also shows that with only text condition, the resulting video experiences significant temporal flickering and sudden appearance changes. These outcomes validate that  when training temporal modules only, the additional image cross attention provides effective visual signals, thereby allowing the temporal module to focus on video consistency, leading to reduced flickering and improved video quality.

\noindent\textbf{Impact of image condition and masked condition on image animation.} Tab.~\ref{tabel:ablation3} shows that masked condition helps to produces the first animation frame that matches the conditioning image, yet cannot guarantee the temporal consistency, as suggested by a high DINO (First) score and low DINO (Avg) score.
Adding the image condition improves temporal consistence and subject identify,
as evidenced by the high DINO (Avg) score and confirmed in Fig.~\ref{fig:ablation3}.
With both masked condition  and image condition, our model produces the first frame that highly preserves the conditioning image, as well as visually consistent animations.
\begin{table}[t]
\centering

\resizebox{0.85\linewidth}{!}{
\begin{tabular}{lcccc}
\toprule
& composed & composed-fs & decomposed-fs & MVB \\
\hline
rate & 0\% & 13\% & 97\% & 97\% \\
\bottomrule
\end{tabular}}
\caption{Image controlent successful rates with different spatial temporal designs in VDMs.}
\label{table:ablation1}
\end{table}
\begin{table}[t]
\resizebox{0.98\linewidth}{!}{
\begin{tabular}{llllll}
\toprule
                           &                & FVD($\downarrow$) & T-V alignment & Motion-Fidelty & Quality \\
                           \hline
\multirow{2}{*}{Text Only} & Ours w/o image condition     & 602 & 8\%           & 4\%            & 0\%     \\
                           & Animatediff-XL & 589 & 40\%          & 12\%           & 30\%    \\
                           \hline
MVB                & Ours           & 542 & 52\%          & 84\%           & 70\% \\  
\bottomrule
\end{tabular}}
\caption{Impact of  multi-modal condition for the VDM.}
\label{table:ablation2}
\end{table}
\begin{table}[t]
\centering
\resizebox{0.85\linewidth}{!}{
\centering
\begin{tabular}{lccc}
\toprule

                 & DINO (First) & DINO (Avg) & CLIP-T (Avg) \\
                 \hline
text only        & 0.264       & 0.262    & 0.285       \\
+ masked condition & 0.760       & 0.296    & 0.210       \\
+ image condtion & 0.638       & 0.562    & 0.282       \\
+ both           & 0.765       & 0.614    & 0.284 \\
\bottomrule
\end{tabular}}
\caption{Impact of conditions masks and decoupled cross attention for image animation. Higher metrics are better. }
\label{tabel:ablation3}
\end{table}
\section{Conclusion}
We present \name, a new video generation model that conditions on both image and text inputs using the Multimodal Video Block (MVB). Our model stands out in producing high-quality videos with controllable visual appearance. In the meantime, our model is able to harvest the pre-trained image ControlNet to control over geometry without extra training overhead.
The model offers a generic architecture and versatile conditioning mechanisms, which allows it to be easily adapted for various video generation tasks, such as image animation, video editing and subject-customized video generation, with superior generation quality than previous methods, revealing its great potential to serve as a foundation model for video generation research and application.

\section{Ethic}
Our model operates with both image and text inputs, and similarly to the Blip-Diffusion~\cite{blipdiffusion}, we aim to ensure a strong relevance of the generation samples to the supplied image. This image can originate from two sources: either directly uploaded by human users or generated by a text-to-image model like DALL-E 3 or Stable Diffusion XL. When the image is produced by a text-to-image model, the bias, fairness, and toxicity levels of our model are inherent and comparable to those of the text-to-image model. Conversely, when the image is provided by the user, these aspects are mostly governed by the user, not our model. To mitigate harmful content that may result from inappropriate image/text inputs, we plan to implement an NSFW (Not Safe For Work) detector, which is open-sourced in diffusers library~\cite{von-platen-etal-2022-diffusers} and can be directly integrated into our model. This detector will strengthen the model governance layer to control the input contents, either from human users or from the text-to-image model, thereby reducing harmful components. 
In addition to the above measures, we are conducting red-teaming experiments to understand and minimize harmful generation contents as a result of red-teaming attacks. 
Note that by default, our model is designed to generate non-toxic elements as long as the provided inputs do not elicit harmful or biased content. 
Regardless, we are committed to eliminating any toxicity or biases in our model and we will not open source or make our model publicly available until the safety measures are properly in place.



{
    \small
    \bibliographystyle{ieeenat_fullname}
    \bibliography{main}
}


\end{document}


\maketitle
\section{Video Results.}

We include our video results and comparisons at \url{https://anonymous-cvpr24-6298.github.io/}. To show that there have been no changes to our website post-deadline, we've added a video titled 'demonstration.mp4' which covers all the website's content. \textbf{Should you encounter issues with video play on the online website, we have included a local version of the website in the supplementary materials. To access it, please click on the file named \textit{"index.html"} and open it with a browser on your local machine.}
\section{Implementation details}
\begin{table}
\centering
\resizebox{0.8\linewidth}{!}{
\begin{tabular}{lc}
\toprule
\textbf{Hyperparameter} & \textbf{Keyframe Module} \\
\toprule
$fps$ & 8 \\
Frames & 16\\
Channels & 320 \\
Depth & 3 \\
Channel multiplier & 1,2,4 \\
Head channels & 64 \\
Image cross attention dim & 2048\\
Text cross attention dim & 2048\\
\midrule
\textit{Training} & \\
Parameterization & $\boldsymbol{\varepsilon}$ \\
\# train steps & 200K \\
Learning rate & $5 \times 10^{-5}$ \\
Batch size per GPU & 1 \\
\# GPUs & 16 \\
GPU-type & A100-40GB \\
$text_{\text{\footnotesize{drop}}}$ & 0.1 \\
$image_{\text{\footnotesize{drop}}}$ & 0.25 \\
\midrule
\textbf{Diffusion Setup} & \\
Diffusion steps & 1000 \\
Noise schedule & Linear \\
$\beta_{0}$ & $0.00085$ \\
$\beta_{T}$ & 0.012 \\
\midrule
\textbf{Sampling Parameters} & \\
Sampler & Euler Discrete \\
Steps & 75 \\
$\eta$ & 1.0 \\
\toprule
\end{tabular}}
\caption{Hyperparameters Details.}
\label{detail}
\end{table}
Initially, we train the  image cross-attention layer while keeping the other weights of SDXL fixed. During this phase, we introduce randomness by omitting the image condition with a 25\% probability and the text condition with a 10\% probability. When it comes to video training, we maintain all spatial parameters, including the weights of SDXL and the image cross-attention layers, unchanged. For an elaborate explanation of our video model, please refer to Table~\ref{detail}.
\section{More ablation studies}
\noindent\textbf{Impact of multimodal condition  on video length for the foundation VDM role. }
\begin{table}[t]
\resizebox{1.0\linewidth}{!}{
\begin{tabular}{lllll}
\toprule
                          & FVD($\downarrow$) & T-V alignment & Motion-Fidelty & Quality \\
                           \hline
 w/o image condition     & 576 / 602 &  44\% / 8\%           & 45\% / 6\%            & 45\% /  3\%     \\

                           \hline
with mulitmodal condition                    & 539 / 542 &  56\% / 92\%          & 55\% / 94\%           &  55\% /97\% \\  
\bottomrule
\end{tabular}}
\caption{Impact of video length: 8 frames / 16 frames and multi-modal condition.}
\label{table:supp2}
\end{table}
We explore the impact of multimodal condtion on different video length as shown in Tab.~\ref{table:supp2} on MSR-VTT dataset. The findings indicate that for shorter videos of 8 frames, text-only conditioning performs comparably to multimodal conditioning. For longer videos of 16 frames, however, multimodal conditioning significantly  surpasses text-only conditioning. This outcome suggests that the additional visual cues from the decoupled multimodal cross-attention layers enhance the temporal layers' focus on long-term consistency, resulting in smoother and higher-quality videos.

\begin{figure}[t]
  \centering
 
   \includegraphics[width=1\linewidth]{figures/supp.pdf}
\vspace{-6mm}
   \caption{Comparisons with Stable Video Diffusion~\cite{boss}.}
   \label{fig:supp}
\end{figure}

\section{More discussion with related and concurrent works}
\subsection{Comparisons with other foundation VDMs}
\subsubsection{Stable Video Diffusion~\cite{boss}}

Our work is concurrent to the Stable Video Diffusion~\cite{boss}(Nov. 22th, 2023), which introduces a text-to-video VDM (Fig.~\ref{fig:supp}(a)) and an image-to-video(image animation) VDM (Fig.~\ref{fig:supp}(b)). These models rely solely on either image or text inputs. The text-only approach struggles with zero-shot subject-specific video generation, while the image-only model finds it challenging to adhere to text directives in animation, as shown in our image animation comparisons part of the  provided  anonymous website. In contrast, as shown in Fig.~\ref{fig:supp}(c),  our method employs decoupled multimodal cross-attention, which effortlessly facilitates zero-shot subject-specific video creation and animates images in line with textual descriptions. 
\textbf{Moreover}, the image ControlNet  is not directly applicable to Stable Video Diffusionn. On the contrary, our model can directly employ the image ControlNet for structural control.
\subsubsection{DynamicCrafter~\cite{xing2023dynamicrafter}}
Our method is concurrent with DynamicCrafter(Oct. 2023), which is designed for animating single images. In contrast, our method serves as a foundational Video Diffusion Model (VDM) capable of performing zero-shot customized video generation, image animation, and video editing tasks. Additionally, our method demonstrates enhanced and better preservation of identity than DynamicCrafter  in image animation tasks, as evidenced in Figure 7 of the main paper and on our website.

\subsubsection{VideoComposer~\cite{videocomposer} }

VideoComposer can manage process control through its video ControlNet, but this necessitates additional training after the primary VDM training, incurring substantial computational costs. In contrast, our approach requires only the training of the base VDM and can directly integrate the image ControlNet without extra training expenses. Additionally, while VideoComposer uses an appearance ControlNet and a single text condition cross-attention layer for image animation, which may lead to abrupt changes and distortions as demonstrated in Figure 7 of the main paper, our method employs a decoupled multimodal cross-attention layer that results in a smoother animation.

\subsubsection{Any to Any Generation~\cite{tang2023any} }
Any to Any Generation enables control across multiple modalities; however, it necessitates a distinct UNet for each modality and a compositional diffusion process involving multiple UNets. Conversely, our method employs a single UNet to achieve multimodal control, facilitated by decoupled multimodal cross-attention layers.

\subsection{Comparisons with IP-Adapter~\cite{ye2023ip-adapter}}IP-Adapter primarily targets image-text co-control within the image domain, whereas our work is centered around the video domain. Our paper highlights three distinct aspects in the video domain with a focus on multimodal conditioning:
\textbf{(1)}We investigate the integration of temporal modules into the Video Diffusion Model (VDM) to enable the direct use of the image ControlNet for video.
\textbf{(2)}We observe that relying solely on text conditions, which lack adequate visual cues, forces temporal modules to compensate by learning additional spatial information. This detracts from their ability to maintain temporal consistency, leading to increased flickering and reduced video quality. However, with the supplementary visual signals provided by decoupled multimodal cross-attention layers, the focus of temporal modules shifts more towards temporal consistency, resulting in smoother and higher-quality videos.
\textbf{(3)}The decoupled multimodal cross-attention layers are particularly beneficial for image animation, as they mitigate abrupt changes in appearance, thereby ensuring smoother animations.

\subsection{Comparisons with other video editing methods.}

\subsubsection{Make-A-Protagonist~\cite{zhao2023makeaprotagonist} } \textbf{(1)} Make-A-Protagonist is exclusively focused on video editing tasks. It necessitates fine-tuning for each new video to accomplish the editing process. In contrast, our approach is designed for broad-spectrum video generation and editing, capable of creating a new video or editing an existing one without any additional training.

\textbf{(2)} Make-A-Protagonist adds the encoded clip image embedding directly into the latent features. Conversely, our approach employs a set of distinct multi-modal cross-attention layers, treating the image features as the value and key for extraction.

\subsubsection{Gen-1~\cite{esser2023structure}}

Gen-1 achieves structural control by concatenating a depth map with the noisy latent input for the UNet. Absent this depth map, the UNet would fail. In contrast, our UNet functions effectively without such structural conditions and can generate high-quality videos.

Additionally, while Gen-1 can control subject appearance, it add text and image embeddings directly together outside the UNet. This method significantly alters the appearance in the image embedding influenced by the text embedding. Conversely, we utilize decoupled multimodal cross-attention layers within the UNet, treating text and image embeddings as separate key and value inputs in the attention layers. This approach allows our method to adhere to text descriptions while simultaneously preserving the image's appearance.

\subsubsection{CCEdit~\cite{feng2023ccedit}.}

CCEdit mandates the utilization of both a base UNet and ControlNet signals during its training phase. In the inference stage, the absence of this depth map results in the UNet's failure. In contrast, our method requires only the base UNet during training and seamlessly integrates the image ControlNet during inference. Moreover, our UNet operates effectively even without ControlNet conditions, enabling the generation of high-quality videos.

{
    \small
    \bibliographystyle{ieeenat_fullname}
    \bibliography{main}
}